\documentclass{article}

\PassOptionsToPackage{numbers, compress}{natbib}
\usepackage[preprint]{neurips_2025}
\usepackage{colortbl}
\usepackage{enumitem}
\usepackage{wrapfig}
\usepackage{graphicx} 
\usepackage{booktabs}
\usepackage{multirow}

\usepackage[utf8]{inputenc} 
\usepackage[T1]{fontenc}    
\usepackage{hyperref}       
\usepackage{url}            
\usepackage{booktabs}       
\usepackage{amsfonts}       
\usepackage{nicefrac}       
\usepackage{microtype}      
\usepackage{xcolor}         
\usepackage{algorithm}
\usepackage{amsmath} 
\usepackage{amsthm}
\usepackage{algorithm}
\usepackage{algorithmic}
\usepackage{multirow}
\usepackage{multicol}
\usepackage{subfigure}

\title{Improving Out-of-Domain Robustness with Targeted Augmentation in Frequency and Pixel Spaces}

\author{%
  Ruoqi Wang \thanks{rwang280@connect.hkust-gz.edu.cn} \\
  HKUST(GZ) \\
  \And
  Haitao Wang \\
  SYSU \\
  \AND
  Shaojie Guo \\
  ECNU \\
  \And
  Qiong Luo \\
  HKUST(GZ)\&HKUST \\
}

\begin{document}

\maketitle

\begin{abstract}

Out-of-domain (OOD) robustness under domain adaptation settings—where labeled source data and unlabeled target data come from different distributions—is a key challenge in real-world applications. A common approach to improving OOD robustness is through data augmentations. However, in real-world scenarios, models trained with generic augmentations can only improve marginally when generalized under distribution shifts toward unlabeled target domains. While dataset-specific targeted augmentations can address this issue, they typically require expert knowledge and extensive prior data analysis to identify the nature of the datasets and domain shift.
To address these challenges, we propose \textbf{Frequency-Pixel Connect}, a domain-adaptation framework that enhances OOD robustness by introducing a targeted augmentation in both the frequency space and pixel space. 
Specifically, we mix the amplitude spectrum and pixel content of a source image and a target image to generate augmented samples that introduce domain diversity while preserving the semantic structure of the source image.
Unlike previous targeted augmentation methods that are both dataset-specific and limited to the pixel space, Frequency-Pixel Connect is dataset-agnostic, enabling broader and more flexible applicability beyond natural image datasets.
We further analyze the effectiveness of Frequency-Pixel Connect by evaluating the performance of our method connecting same-class cross-domain samples while separating different-class examples. 
We demonstrate that Frequency-Pixel Connect significantly improves cross-domain connectivity and outperforms previous generic methods on four diverse real-world benchmarks across vision (+3.0\%), medical (+4.9\%), audio (+6.4\%), and astronomical domains (+9.1\%), and it also outperforms other dataset-specific targeted augmentation methods.
\end{abstract}

\section{Introduction}

In real-world applications, the distributions in the training datasets often differ from those in deployment environments \cite{koh2021wilds, qu2024connect}. In such out-of-distribution (OOD) settings, models trained on source domains typically degrade when applied to unlabeled target domains. For example, in wildlife monitoring, ecologists use models to classify species from camera trap images, but these models often experience substantial accuracy drops when applied to new locations \cite{beery2018recognition}. Annotated data are usually available only for a limited subset of cameras, which may not capture the diversity of environments in the unlabeled target domain. In this paper, we focus on improving OOD robustness in domain adaptation, where we have labeled data from source domains and unlabeled data from target domains. 

Data augmentation is a common strategy for improving OOD robustness, but designing effective augmentations remains challenging. Generic augmentations of images (e.g., RandAugment\cite{cubuk2020randaugment}, CutMix\cite{yun2019cutmix}) have only marginal and inconsistent gains across datasets \cite{gulrajanisearch, wilesfine, hendrycks2021many}, while domain-invariance methods \cite{yan2020improve, zhou2020deep, ilse2021selecting, gulrajanisearch, yao2022improving} have limited improvements, especially on real-world domain shifts \cite{gao2023out}. Although dataset-specific augmentation strategies \cite{gao2023out, qu2024connect} have demonstrated promising improvements in OOD performance, they typically require expert domain knowledge and extensive preliminary analysis, limiting their scalability and general applicability. For example, in wildlife recognition tasks, these methods randomize background regions in camera trap images to mitigate spurious correlations with location-specific terrains or lighting conditions, thereby encouraging the model to focus on the animal foreground. However, such augmentations depend on the availability of segmentation masks, restricting their transferability to datasets without foreground annotations. Similarly, in tumor identification from histopathology slides, stain color variations across hospitals are addressed through Stain Color Jitter \cite{tellez2018whole}, which perturbs images within the hematoxylin and eosin staining space to suppress color-related domain biases while preserving morphological features \cite{gao2023out}. Nonetheless, this approach relies on specialized knowledge of staining properties and color space manipulations, and is not adaptable to datasets where domain shifts arise from different sources. These limitations highlight the need for more general, dataset-agnostic augmentation strategies to improve OOD robustness across diverse real-world settings.


To address these challenges, we propose Frequency-Pixel Connect, a generic and dataset-agnostic domain adaptation framework that introduces targeted augmentations in both the frequency and pixel spaces. The training process consists of two stages: pretraining with generic augmentations to learn domain-specific representations, followed by fine-tuning with our generic, dataset-agnostic targeted augmentation in both the frequency space and pixel space to enhance cross-domain robustness. Unlike previous targeted augmentation methods that rely on dataset-specific priors and operate solely in pixel space, Frequency-Pixel Connect automatically simulates domain shifts without requiring expert knowledge or manual analysis. Specifically, given two images from different domains, we perform linear interpolation between their amplitude spectra to create frequency-space augmentations that preserve the semantic structure of the source image while introducing domain-diverse signals. Meanwhile, we apply pixel-space blending to add complementary spatial information in the augmentations. The final augmented image is obtained by fusing these two results from the frequency space and the pixel space. 

We evaluate our method on four real-world benchmarks: wildlife recognition (iWildCam \cite{beery2021iwildcam, sagawaextending}), tumor detection (Camelyon17 \cite{bandi2018detection, sagawaextending}), bird species recognition (BirdCalls \cite{joly2022overview, gao2023out}), and galaxy morphology classification (Galaxy10 DECaLS \& SDSS \cite{Galaxy10}). Extensive experiments demonstrate that Frequency-Pixel Connect consistently outperforms other methods. Compared to other generic methods, Frequency-Pixel Connect achieves notable improvements in OOD performance, including +3.0\% F1-score on iWildCam, +4.9\% accuracy on Camelyon17, +6.4\% F1-score on BirdCalls, and +9.1\% accuracy on Galaxy10. Also, it  surpasses dataset-specific augmentation strategies tailored for different datasets. We further analyze the effectiveness of Frequency-Pixel Connect by evaluating the performance of our method connecting same-class cross-domain samples while separating different-class examples.

Our work makes the following key contributions: (1) Frequency-Pixel Connect is the first work to introduce targeted augmentations in both frequency and pixel spaces to improve out-of-domain (OOD) robustness during fine-tuning. (2) Our method is generic and broadly applicable, demonstrating strong performance across diverse real-world applications without needing dataset-specific adaptation techniques. (3) Frequency-Pixel Connect consistently outperforms other generic baselines, achieving up to +9.1\% OOD accuracy improvement (Galaxy10) and significant gains in other benchmarks, including iWildCam, Camelyon17, BirdCalls, and Galaxy10.

\section{Problem Setting}

\subsection{Feature Decomposition}
\label{Feature Decomposition}
To better understand how models behave under domain shifts, we adopt the feature decomposition framework \cite{gao2023out}, which categorizes input features based on their dependence on the label and domain. Specifically, input features can be decomposed into four types based on their dependence on the label and the domain: 
(1) features that are \textit{label-dependent} and \textit{domain-independent} ($x_\text{obj}$); 
(2) \textit{label-dependent} and \textit{domain-dependent} features ($x_\text{d:robust}$); 
(3) \textit{label-independent} and \textit{domain-dependent} features ($x_\text{d:spu}$); and 
(4) \textit{label-independent} and \textit{domain-independent} features ($x_\text{noise}$).
 We formalize these relationships via their (in)dependence with the label $y$ and domain $d$:
\[
x_\text{obj},\; x_\text{d:robust} \not\!\perp\!\!\!\perp\; y, \quad
x_\text{noise},\; x_\text{d:spu} \perp\!\!\!\perp\; y, \quad
x_\text{d:robust},\; x_\text{d:spu} \not\!\perp\!\!\!\perp\; d, \quad
x_\text{obj},\; x_\text{noise} \perp\!\!\!\perp\; d.
\]
Models should focus on both $x_\text{obj}$ and $x_\text{d:robust}$ while avoiding reliance on spurious or noisy features \cite{gao2023out}. Our connectivity-based analysis and augmentation strategy are motivated by this decomposition.

\begin{figure}[t]
\centerline{\includegraphics[height=6.3cm]{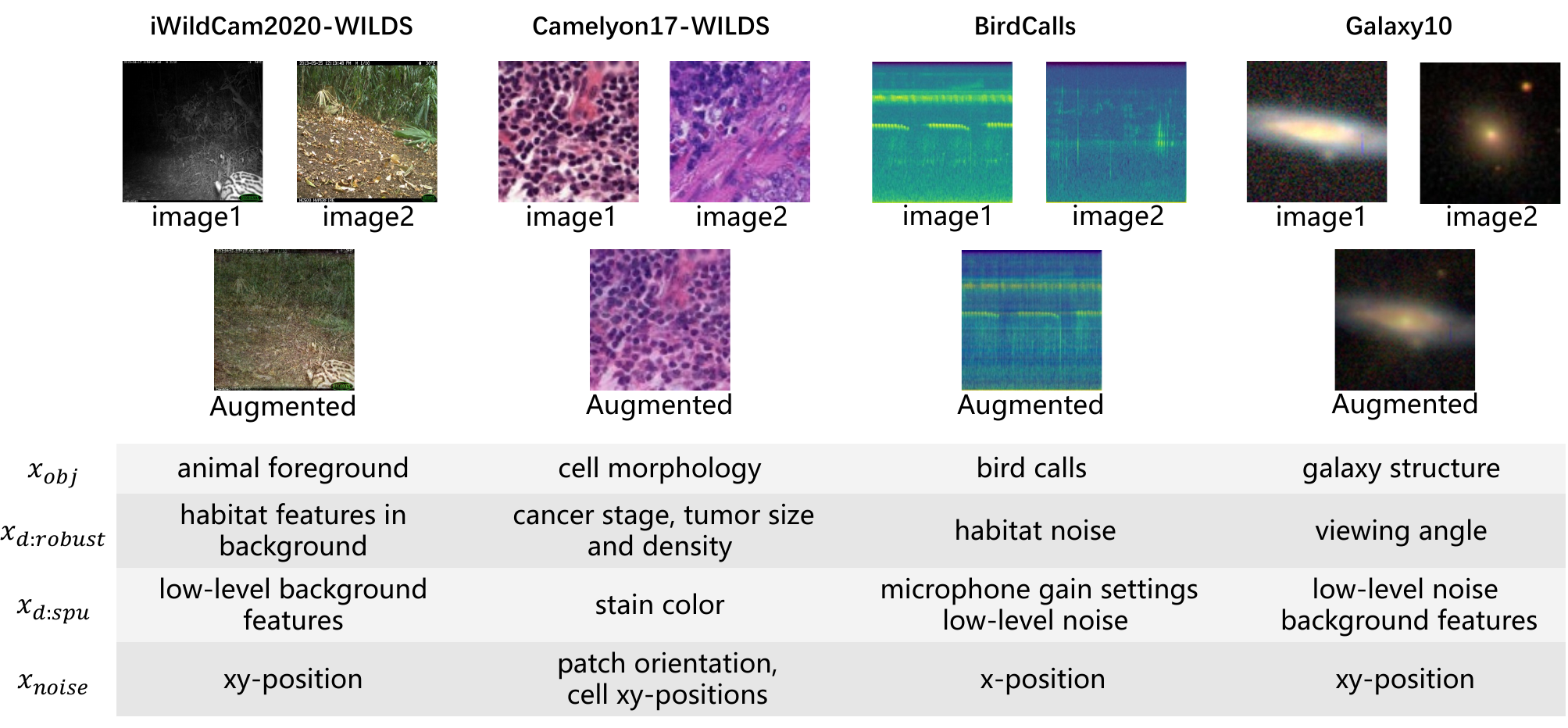}}
\caption{
\textbf{Feature decomposition and augmentation examples.} 
\textit{Top:} For each dataset (iWildCam, Camelyon17, BirdCalls, Galaxy10), we show a source image, its corresponding augmented image generated by our method.
\textit{Bottom:} Based on the decomposition framework \cite{shen2022connect}, we annotate representative features in the datasets across $x_{\text{obj}}$, $x_{d\text{:robust}}$, $x_{d\text{:spu}}$, $x_{\text{noise}}$. 
Our method effectively randomizes $x_{d\text{:spu}}$, varies $x_{d\text{:robust}}$ while preserving or  $x_{\text{obj}}$.
}
\label{4x}
\end{figure}

\subsection{Real-world Datasets}
As presented in Figure \ref{4x}, we evaluate our method on four real-world datasets in this paper. \textbf{Species Classification (iWildCam).}
The iWildCam dataset~\cite{beery2021iwildcam, koh2021wilds} involves classifying animal species \(y\) from images \(x\) taken by static camera traps at 243 locations. While foreground animals contain label-relevant features \(x_{\text{obj}}\), habitat features \(x_{d:\text{robust}}\) vary across locations (e.g., Kenyan savannas vs. Guatemalan forests) and must be disentangled for robust prediction. \textbf{Tumor Identification (Camelyon17).}
Camelyon17~\cite{beery2021iwildcam, koh2021wilds} contains histopathology slides from three hospitals, with domain-specific staining and cancer stage distributions. Domain shifts affect features like color (\(x_{d:\text{spu}}\)) and tumor density (\(x_{d:\text{robust}}\)), making generalization challenging. Robust models must extract diagnostic features \(x_{d:\text{robust}}\) while ignoring spurious variations. \textbf{Bird Species Recognition (BirdCalls).}
BirdCalls~\cite{joly2022overview, gao2023out} includes audio clips from nine microphones. Species identity \(y\) is conveyed by bird calls, and domain-specific noise stems from habitat (\(x_{d:\text{robust}}\)) and hardware differences (\(x_{d:\text{spu}}\)). Effective models should focus on species cues while randomizing spurious domain noise. \textbf{Galaxy Morphology Classification (Galaxy10).}
Galaxy10~\cite{Galaxy10} comprises DECaLS and SDSS subsets from two different telescopes, differing in resolution and color profile. Galaxy morphology is classified by the main structure of galaxies. \(x_{d:\text{robust}}\) includes viewing angle and \(x_{d:\text{spu}}\) includes noise and background features. A more detailed introduction of the four datasets is presented in Appendix Section \ref{Dataset Details}.


\subsection{Connectivity}

To better understand how data augmentations impact cross-domain generalization, we adopt the concept of \textit{connectivity} \cite{shen2022connect}. 
We define connectivity between a class-domain pair $((y_1, d_1), (y_2, d_2))$ under four conditions:

\begin{equation}
\left\{
\begin{aligned}
\rho &: y_1 = y_2,\ d_1 = d_2 \quad &\text{(same class, same domain)} \\
\alpha &: y_1 = y_2,\ d_1 \ne d_2 \quad &\text{(same class, different domain)} \\
\beta &: y_1 \ne y_2,\ d_1 = d_2 \quad &\text{(different class, same domain)} \\
\gamma &: y_1 \ne y_2,\ d_1 \ne d_2 \quad &\text{(different class, different domain)}
\end{aligned}
\right.
\end{equation}

Here, each value indicates how “connected” the pairs are. Shen et al. \cite{shen2022connect} show that the ratios $\frac{\alpha}{\gamma}$ and $\frac{\beta}{\gamma}$ exhibit strong empirical correlation with OOD accuracy.  In this work, we use this insight to analyze how our augmentation changes the connectivity of the data. More details on the computation of connectivity are presented in Section \ref{Empirical Evaluations of Connectivity} and the Technical Appendix Section \ref{appendix_connectivity}.

\subsection{Motivation}

\begin{wrapfigure}{r}{0.55\textwidth}
\centering
\includegraphics[width=0.55\textwidth]{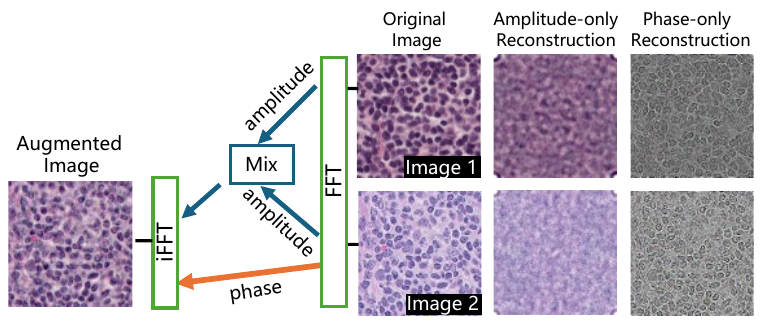}
\caption{The figure shows the amplitude-only and phase-only reconstructions of the two input images, and the process where the amplitudes of Image 1 and Image 2 are first mixed, and the mixed amplitude is combined with the phase of Image 2 to generate an augmentation of Image 2.}
\label{frequency}
\end{wrapfigure}

\begin{figure}[t]
\centerline{\includegraphics[height=5cm]{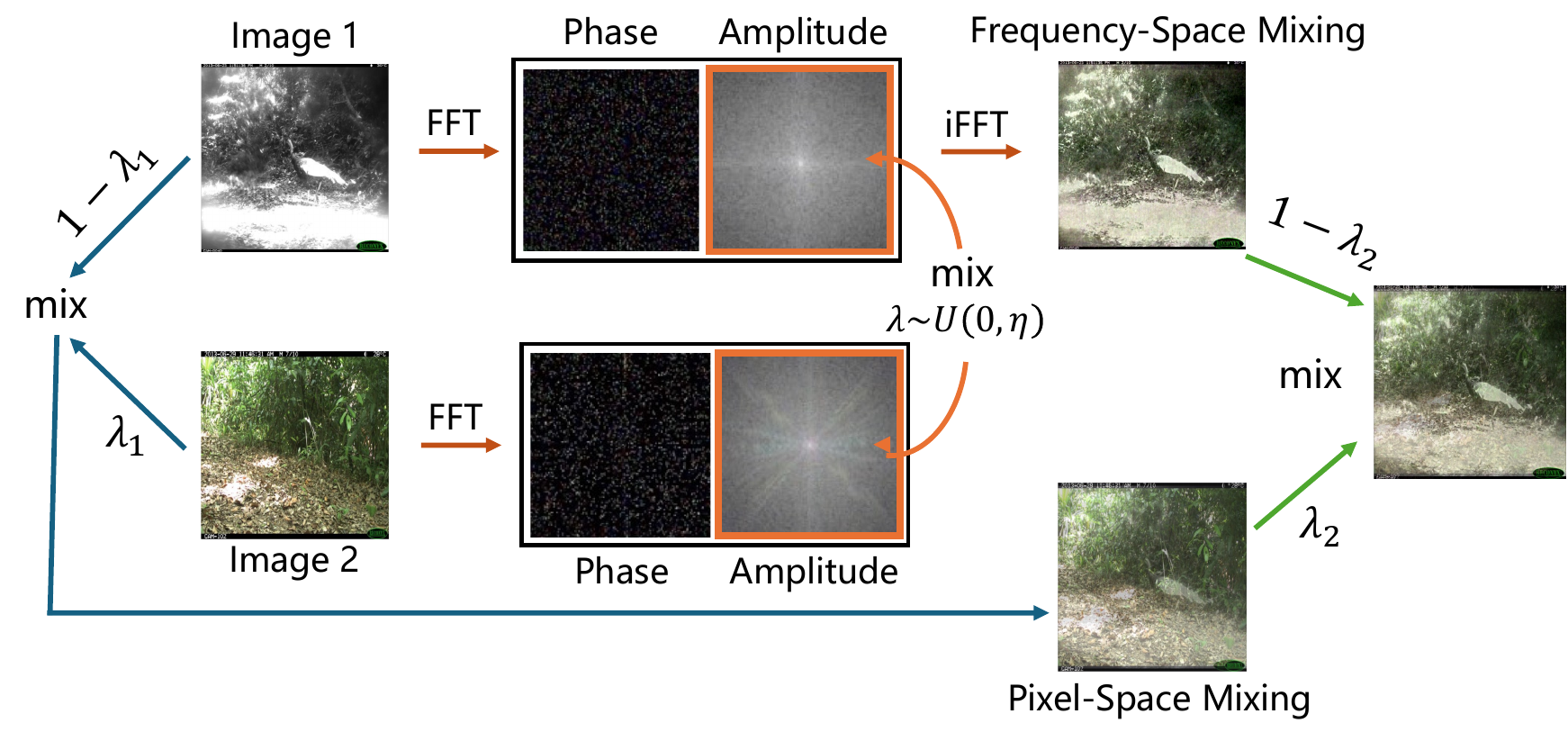}}
\caption{Overview of our Frequency-Pixel Mixing strategy. Given two images, we first do amplitude interpolation in the frequency space and pixel-space blending with mixing ratio $\lambda_1$. The final augmented image is obtained by combining the two intermediate results with a mixing ratio $\lambda_2$.}
\label{ouraug}
\end{figure}

Our method is motivated by observing and analyzing how frequency and pixel-space mixing affect different feature components under domain shift. 
Out-of-domain (OOD) adaptation requires perturbing domain-specific features while preserving task-relevant semantics. From the feature decomposition perspective~\cite{shen2022connect}, this means randomizing spurious domain-dependent components ($x_{d\text{:spu}}$), and preserving label-relevant content ($x_{\text{obj}}$ and $x_{d\text{:robust}}$). As in Figure \ref{frequency}, we observe that frequency-based amplitude mixing (Details of the mixing strategy are provided in Section \ref{Method}.) preserves task-relevant semantic structure \cite{xu2023fourier} ($x_{\text{obj}}$ and $x_{d\text{:robust}}$), while effectively randomizing $x_{d\text{:spu}}$ (Another example is the \textit{Frequency-Space Mixing} in Figure \ref{ouraug}). However, excessive frequency mixing may introduce artifacts, influencing the extraction of detailed features. 

\begin{wraptable}{r}{6.9cm}
\centering
\caption{Comparison of connectivity values and ratio on iWildCam dataset.}
\begin{tabular}{l|ccc|cc}
\toprule
\textbf{Aug} & $\alpha$ & $\beta$ & $\gamma$ & $\alpha/\gamma$ & $\beta/\gamma$ \\
\midrule
w/o & 0.007 & 0.008 & 0.021 & 0.33 & 0.38 \\
Ours     & 0.212 & 0.030 & 0.051 & 4.16 & 0.59 \\
\bottomrule
\end{tabular}
\label{connect_test}
\end{wraptable}

To complement this, we introduce pixel-space mixing. By spatial blending of pixel values, pixel-space mixing brings back more pixel-level details and perturbs $x_{d\text{:spu}}$. However, it may influence $x_{\text{obj}}$ and $x_{d\text{:robust}}$ as well. To address this problem, we then define a flexible tunable ratio (details are in Section \ref{Method} and \ref{Parameters}). By combining these two complementary perturbation paths, our goal is to balance the label-relevant and domain-relevant shifts, achieving domain diversity while maintaining semantic identity. This dual-space augmentation enables fine-grained control over domain perturbation intensity and promotes robust feature learning across source and unseen domains. 
We then empirically evaluate this intuitive proposal by comparing the connectivity ratios $\frac{\alpha}{\gamma}$ and $\frac{\beta}{\gamma}$ of the augmented and unaugmented datasets. Results in Table \ref{connect_test} show that our proposed augmentation can enhance the cross-domain connectivity, achieving domain diversity while maintaining label-relevant identity.

\section{Method}
\label{Method}
\textbf{Frequency-Pixel Connect} uses a pretrain-finetuning process, where a model pretrained with generic augmentations is then fine-tuned using targeted augmentations. Instead of requiring specialized augmentations for different datasets, we introduce a \textbf{Frequency-Pixel Mixing} augmentation strategy, which operates in both Fourier (or frequency) and pixel spaces to generate effective augmentations.
 
\subsection{Frequency-Pixel Mixing}

For an image $x$, its Fourier transformation $\mathcal{F}(x)$ is formulated as:
\begin{equation}
\mathcal{F}(x)(u, v) = \sum_{h=0}^{H-1} \sum_{w=0}^{W-1} x(h, w) e^{-j2\pi\left( \frac{hu}{H} + \frac{wv}{W} \right)},
\end{equation}
where $(u,v)$ are coordinates in the frequency space. $\mathcal{F}^{-1}(x)$ denotes the corresponding inverse Fourier transformation. Both forward and inverse Fourier transforms can be efficiently computed via the Fast Fourier Transform (FFT) algorithm.

The amplitude and phase components of $\mathcal{F}(x)$ are computed respectively as:
\begin{equation}
\mathcal{A}(x)(u, v) = \left[ R^2(x)(u,v) + I^2(x)(u,v) \right]^{1/2},
\quad
\mathcal{P}(x)(u, v) = \arctan\left( \frac{I(x)(u,v)}{R(x)(u,v)} \right),
\end{equation}
where $R(x)$ and $I(x)$ denote the real and imaginary parts of $\mathcal{F}(x)$, respectively. For RGB images, the Fourier transform is applied independently to each channel.

\noindent
\textbf{Frequency-based Augmentation:}  
Given two source images $x_1$ and $x_2$, we aim to generate the augmentation of image $x_1$ using the amplitude of $x_2$. We linearly interpolate between the amplitude spectrums of two images from arbitrary source domains \cite{xu2023fourier}. Specifically, we randomly select a square crop of ratio $r$ in the amplitude spectra and interpolate between $x_1$ and $x_2$:
\begin{equation}
\mathcal{A}(\hat{x})(u, v) = (1-\lambda) \mathcal{A}(x_1)(u,v) + \lambda \mathcal{A}(x_2)(u,v),
\end{equation}
where $\lambda \sim U(0, \eta)$ is sampled from a uniform distribution controlling the mixing intensity, and the interpolation is only applied within the cropped region.

The new Fourier representation is constructed by combining the interpolated amplitude and the original phase of $x_1$:
\begin{equation}
\mathcal{F}(\hat{x})(u, v) = \mathcal{A}(\hat{x})(u, v)  e^{-j\mathcal{P}(x_1)(u, v)}.
\end{equation}
The augmented image $\hat{x}_f$ is then obtained by applying the inverse Fourier transform:
\begin{equation}
\hat{x}_f = \mathcal{F}^{-1}(\mathcal{F}(\hat{x})).
\end{equation}

\noindent
\textbf{Pixel-space Mixing:}  
In addition to frequency-domain augmentation, we introduce an additional pixel-space mixing strategy to add more pixel-level details in the augmentation. Specifically, given an input image $x$ and a background image $x'$, we first apply pixel-wise blending with ratio $\lambda_1$:
\begin{equation}
\hat{x}_p = (1 - \lambda_1)x_1 +\lambda_1 x_2,
\end{equation}

Finally, we combine both pixel-space and frequency-space augmented images via a second-stage blending:
\begin{equation}
\hat{x} = (1 - \lambda_2)\hat{x}_f + \lambda_2 \hat{x}_p,
\end{equation}
This two-stage fusion allows us to perturb both global style (via amplitude spectrum) and local spatial cues, while maintaining the semantic content of the original sample and the pixel details.

\subsection{Training Framework}
\textbf{Pretraining with Generic Augmentations}
Given source and target distributions $P_S$ and $P_T$ over input space $\mathcal{X}$, we study unsupervised domain adaptation. We observe labeled samples $(x, y) \sim P_S$, where $y \in \mathcal{Y}$ are sampled from the label distribution $p^*(\cdot \mid x)$, and unlabeled samples in the target distribution $x \sim P_T$. The mixed unlabeled distribution is denoted by $P_U = \beta P_S + (1 - \beta) P_T$, where $\beta \in [0, 1]$. In this paper, we adopt a contrastive pretraining strategy \cite{qu2024connect} to train a representation encoder $\phi: \mathcal{X} \rightarrow \mathbb{R}^k$ using augmented views of unlabeled inputs. Let $\mathcal{S}_+(x, x^+)$ denote the distribution of positive pairs $(x, x^+)$, where $x$ and $x^+$ are augmentations of a common input $\bar{x}$:
$
\mathcal{S}_+(x, x^+) = \mathbb{E}_{\bar{x} \sim P_U}\left[\mathcal{A}_{\text{pre}}(x \mid \bar{x}) \cdot \mathcal{A}_{\text{pre}}(x^+ \mid \bar{x})\right],
$
where $\mathcal{A}_{\text{pre}}$ denotes the pretraining-time augmentation distribution. We minimize the following contrastive loss to encourage small distance $d_+$ between positive pairs and large distance $d_-$ between negative pairs:
\begin{equation}
\mathcal{L}_{\text{pretrain}}(\phi) = 
\mathbb{E}_{(x, x^+) \sim \mathcal{S}_+} \left[d_+\left(\phi(x), \phi(x^+)\right)\right]
- \mathbb{E}_{x, x' \sim P_U} \left[d_-\left(\phi(x), \phi(x')\right)\right].
\end{equation}

The output of pretraining is a fixed encoder $\hat{\phi}$:
$\hat{\phi} = \arg\min_\phi \mathcal{L}_{\text{pretrain}}(\phi).$

In the fine-tuning stage, we train a prediction head $h: \mathbb{R}^k \rightarrow \mathbb{R}^n$ on top of the frozen encoder $\hat{\phi}$ using labeled source data. Given a labeled dataset $(x, y) \sim P_S$, and applying augmentations $\mathcal{A}_{\text{ft}}$ during fine-tuning, the objective is:
\begin{equation}
\mathcal{L}_{\text{ft}}(h) = 
\mathbb{E}_{x \sim P_S,\, y \sim p^*(\cdot \mid x),\, x' \sim \mathcal{A}_{\text{ft}}(\cdot \mid x)} 
\left[\text{loss}_{\text{ft}}\left(h(\hat{\phi}(x')), y; \theta\right)\right].
\end{equation}

\textbf{Linear Probing then Fine-Tuning.}
We adopt the linear probing then fine-tuning (LP-FT) strategy \cite{kumar2022fine} in our framework, which has been shown to improve ID and OOD performance over vanilla fine-tuning \cite{qu2024connect}. This procedure first trains a linear classifier on frozen pretrained features, followed by joint optimization of the encoder and classifier.

\section{Experiments}
In this section, we first conduct a comparison between our method and a number of existing methods. We also do an ablation study to evaluate the contribution of each module in our approach. Then we empirically evaluate the connectivity of class-domain pairs on iWildCam and Camelyon17 datasets. Finally, we analyze the effect of mixing parameters.

\subsection{Experimental Settings}

\textbf{Platform:}
We conduct all experiments on a server with 128 AMD EPYC 7543 32-Core Processors, 512GB main memory, and eight NVIDIA RTX A6000 GPUs, each with 48GB device memory. The operating system is Ubuntu 22.04. Our model is implemented in PyTorch 2.1.0.

\textbf{Evaluation Metrics:}
Following previous work \cite{gao2023out, qu2024connect, sarkar2024demystifying, wang2023galaxy}, we select metrics that best suit the characteristics of each dataset to ensure a comprehensive evaluation of model performance. We use F1 score and accuracy as evaluation metrics, depending on the nature of the task and dataset. The F1 score is used for iWildCam,  BirdCalls datasets to account for class imbalance, and we use accuracy for Camelyon17 and Galaxy datasets.

\textbf{Method under Comparison:} 
We compare our method against various baseline approaches for domain generalization and out-of-distribution (OOD) performance enhancement. These baselines include: (1) \textbf{Generic Augmentations.}
For image datasets such as iWildCam, Camelyon17 and Galaxy10, we compare our approach with several commonly used augmentation techniques: RandAugment \cite{cubuk2020randaugment}, CutMix \cite{yun2019cutmix}, MixUp \cite{zhang2017mixup}, and Cutout \cite{devries2017improved}. In the case of the BirdCalls dataset, we compare with MixUp, SpecAugment \cite{park2019specaugment}, random low/high pass filters, noise reduction \cite{sainburg2024noisereduce} and Jitter \cite{tellez2018whole,gao2023out}. (2) \textbf{Dataset-Specific Targeted Augmentations.}
We compared our method with human-specialized dataset-specific targeted augmentations \cite{gao2023out}: Copy-Paste for iWildCam, Stain Color Jitter for Camelyon17, and Copy-Paste + Jitter for BirdCalls. We also compare to the latest SOTA method Connect Later, which included those specific targeted augmentations in the finetuning procedure after pretraining \cite{qu2024connect}. (3) \textbf{Domain Invariance Baselines.}
We compare to LISA \cite{yao2022improving}, a method designed to encourage domain invariance by applying mixing augmentations to inputs of the same class across different domains. Additionally, we evaluate other domain invariance algorithms that do not rely on augmentation, including DANN \cite{long2018conditional, ganin2016domain}, DeepCORAL \cite{sun2016deep, sun2017correlation}, and ERM \cite{qu2024connect}, LP-FT \cite{kumar2022fine}.

More details regarding the implementation settings are listed in Appendix Section \ref{Experimental Details}.

\subsection{Results}

\begin{figure}[t]
    \subfigure{\includegraphics[width=0.98\linewidth]{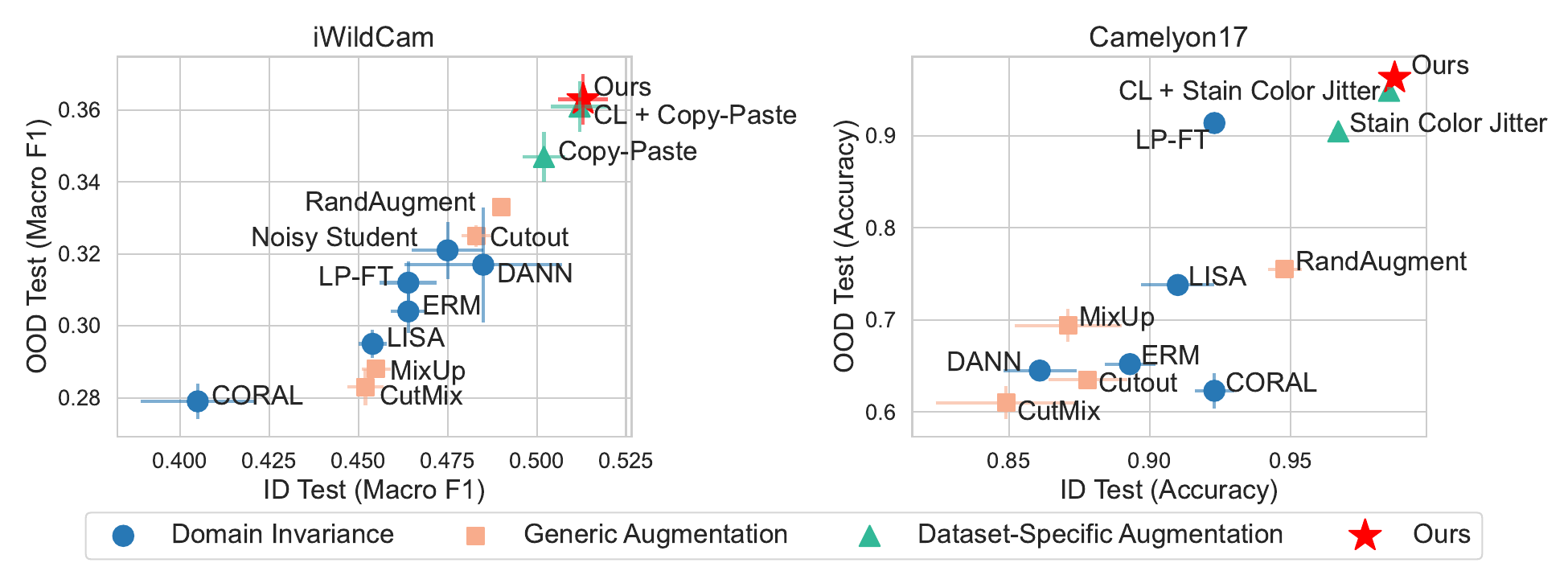}}
  \\

    \subfigure{\includegraphics[width=1\linewidth]{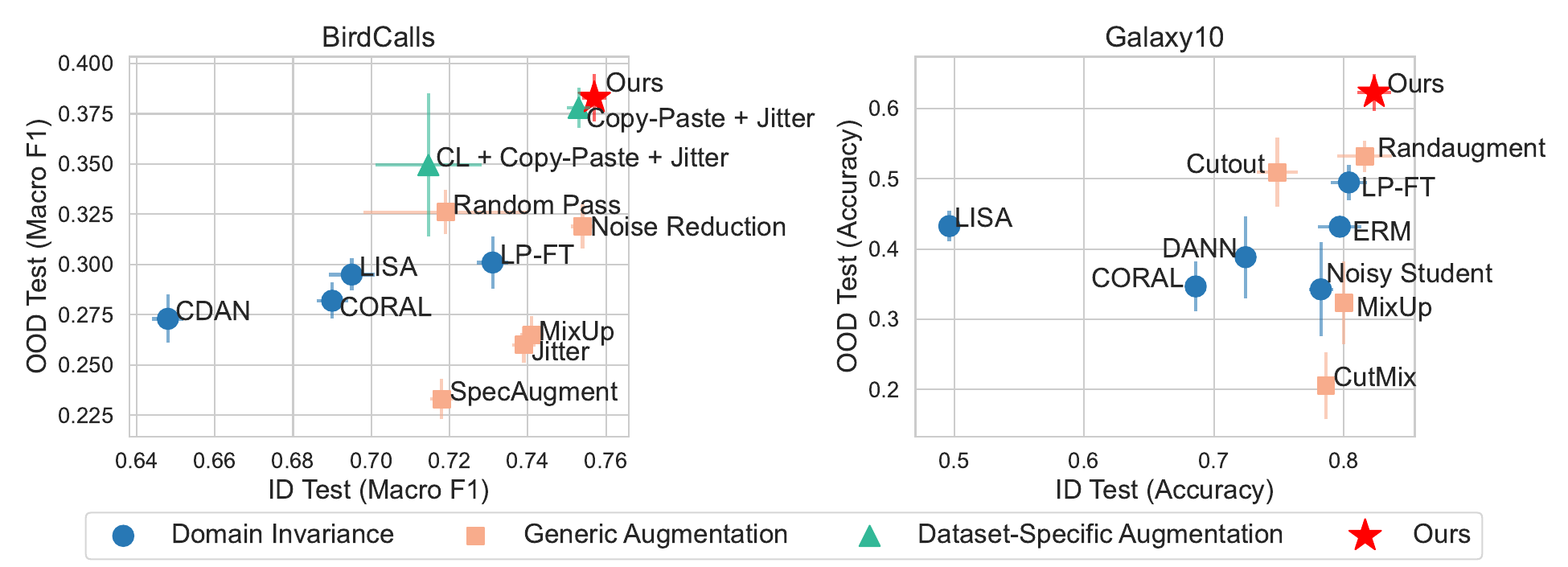}}
  \\
  \caption{We plot the in-domain (ID) performance versus out-of-domain (OOD) performance for all methods across four datasets. Our method consistently outperforms all baselines in OOD generalization. `CL' refers to `Connect Later'.
  }
  \label{points}
\end{figure}

Figure \ref{points} summarizes the trade-off between in-domain (ID) and out-of-domain (OOD) performance for all methods across four datasets: iWildCam, Camelyon17, BirdCalls, and Galaxy10. Each point represents one method, color-coded by method type, and our method is marked as a red star.

Our method consistently achieves the best OOD performance across all datasets, while maintaining strong ID performance. Notably, our approach significantly outperforms all generic methods, including \textbf{generic augmentation} and \textbf{domain invariance} baselines, which do not require dataset-specific design or prior knowledge. Compared to the best generic method (RandAugment) on iWildCam, our method improves OOD Macro F1 from 33.3\% to 36.3\%. On Camelyon17, our approach achieves 96.3\% OOD accuracy, surpassing the best generic augmentation (RandAugment at 75.5\%) and the best domain invariance method (LP-FT at 91.4\%). On BirdCall, OOD Macro F1 increases from 31.9\% (Random Pass) to 38.3\%, and on Galaxy, our method improves OOD accuracy from 53.2\% (Randaugment) to 62.3\%. This demonstrates that our method provides a more effective and generalizable way to improve OOD robustness.

Moreover, our method also surpasses \textbf{dataset-specific augmentation} strategies (Connect Later + ) Copy-Paste on iWildCam, (Connect Later + ) Stain Color Jitter on Camelyon17, and (Connect Later + ) Copy-Paste + Jitter on BirdCalls, all of which require expert knowledge and manual design tailored to the dataset based on the characteristics of the datasets. In contrast, our method operates in a fully \textbf{dataset-agnostic} manner, yet still achieves superior or comparable results. This suggests that our Frequency-Pixel Connect is not only effective but also broadly applicable, without relying on dataset-specific cues or prior knowledge.

\paragraph{Ablation Study.}
Table~\ref{tab:ablation} presents ablation results that isolate the effects of frequency-space and pixel-space mixing. Across all four datasets, our full method consistently achieves the highest OOD performance, outperforming both LP-FT \cite{kumar2022fine} and the single-modality variants. Frequency-only mixing proves more effective than pixel-only, especially for Camelyon17 and Galaxy10, indicating the importance of frequency space perturbation. Importantly, combining frequency and pixel mixing leads to further gains, suggesting that dual-space augmentation provides complementary benefits. The consistent improvement across diverse tasks confirms that our method is both effective and broadly applicable.

\begin{table}[h]
\centering
\caption{
Ablation results comparing LP-FT (Ours without augmentations), pixel-only, frequency-only, and our full method across four datasets. 
}
\small
\begin{tabular}{lcccccccc}
\toprule
& \multicolumn{2}{c}{{iWildCam (F1)}} & \multicolumn{2}{c}{{Camelyon17 (Acc)}} & \multicolumn{2}{c}{{BirdCall (F1)}} & \multicolumn{2}{c}{{Galaxy10 (Acc)}} \\
\cmidrule(lr){2-3} \cmidrule(lr){4-5} \cmidrule(lr){6-7} \cmidrule(lr){8-9}
& {ID} & {OOD} & {ID} & {OOD} & {ID} & {OOD} & {ID} & {OOD} \\
\midrule
LP-FD & 46.4 & 31.2 & 92.3 & 91.4 & 73.1 & 30.1 & 80.4 & 49.5  \\
Pixel-only & 45.4 & 29.1 & 87.5 & 69.8 & 75.2 & 32.4 & 79.6 & 50.5 \\
Frequency-only & \textbf{51.3} & 34.2 & 98.5 & 95.3 & 73.4 & 35.1 & 81.8 & 58.5 \\
Ours & \textbf{51.3} & \textbf{36.3} & \textbf{98.7} & \textbf{96.3} & \textbf{75.7} & \textbf{38.3} & \textbf{82.4} & \textbf{62.3} \\
\bottomrule
\end{tabular}
\label{tab:ablation}
\end{table}

\subsection{Empirical Evaluations of Connectivity}
\label{Empirical Evaluations of Connectivity}
To better understand why our method is effective on improving OOD robustness, we empirically evaluate the connectivity measures for iWildCam and Camelyon17. To compute the average connectivity between two class-domain pairs $(c, d)$ and $(c', d')$, the process involves labeling all training examples from class $c$ and domain $d$ as 0, and those from class $c'$ and domain $d'$ as 1, while discarding examples from other classes or domains. 
Then, a ResNet50 \cite{he2016deep} model is trained in a unified way (Training details are presented in Appendix Section \ref{appendix_connectivity}). 
The test set is prepared similarly to the training set, and the model's performance on test data provides an estimate of connectivity between the specified class-domain pairs, quantified by the test error rate. Specifically, we train binary classifiers from scratch to predict the class-domain pair of each input example. For iWildCam, we randomly select 15 class-domain pairs, while for Camelyon17, we use all class-domain pairs since it is a binary classification task. Our results are presented in Table \ref{connect}.

\begin{table}[h]
\centering
\caption{Comparison of connectivity across class, domain, and both for different methods on Camelyon17 and iWildCam datasets.}
\small
{
\begin{tabular}{l|l|ccc|cc}
\toprule
\textbf{Dataset} & \textbf{Augmentations} & $\alpha$ & $\beta$ & $\gamma$ & $\alpha/\gamma$ & $\beta/\gamma$ \\
\midrule
\multirow{4}{*}{Camelyon17}
& w/o Augmentations & 0.015 & 0.189 & 0.002 & 7.50 & 94.5 \\
& RandAugment        & 0.181 & 0.186 & 0.114 & 1.59 & 1.63  \\
& Stain Color Jitter & 0.029 & 0.116 & 0.004 & 7.25 & 29.0 \\

& Ours      & 0.120 & 0.073 & 0.003 & 40.0 & 24.3 \\
\midrule
\multirow{4}{*}{iWildCam}
& w/o Augmentations & 0.007 & 0.008 & 0.021 & 0.33 & 0.38 \\
& RandAugment        & 0.224 & 0.125 & 0.130 & 1.72 & 0.96 \\
& Copy-Paste         & 0.254 & 0.031 & 0.063 & 4.03 & 0.49 \\
& Ours      & 0.212 & 0.030 & 0.051 & 4.16 & 0.59 \\
\bottomrule
\end{tabular}
}
\label{connect}
\end{table}

To understand the effect of augmentation on feature structure, we report connectivity ratios $\alpha/\gamma$ and $\beta/\gamma$ on Camelyon17, following Shen et al.~\cite{shen2022connect}. 
As shown in Table~\ref{connect}, on both datasets, our method achieves the highest $\alpha/\gamma$, indicating more consistent semantic alignment across domains and effective randomizing spurious domain-dependent components $x_{d\text{:spu}}$. However, moderate $\beta/\gamma$ indicate our method may also mildly perturbs class-relevant features $x_{\text{obj}}, x_{d\text{:robust}}$. However, from Table \ref{tab:ablation} and Figure \ref{points}, our method can still have the best OOD performance.
This result is consistent with the estimation proposed by Shen et al.~\cite{shen2022connect} $
\text{target accuracy} \approx (\alpha/\gamma)^{w_1} \cdot (\beta/\gamma)^{w_2},
$
with higher weight placed on $\alpha/\gamma$ (i.e., $w_1 > w_2$).

These results highlight that our method achieves a desirable balance of the preservation of class-relevant features and the perturbation of domain-relevant features.

\subsection{Effect of Mixing Ratios \texorpdfstring{$\lambda_1$}{lambda_1} and \texorpdfstring{$\lambda_2$}{lambda_1}}
\label{Parameters}

\begin{figure}[h]
    \subfigure[Accuracy Heatmaps on Galaxy10.]{\includegraphics[width=0.5\linewidth]{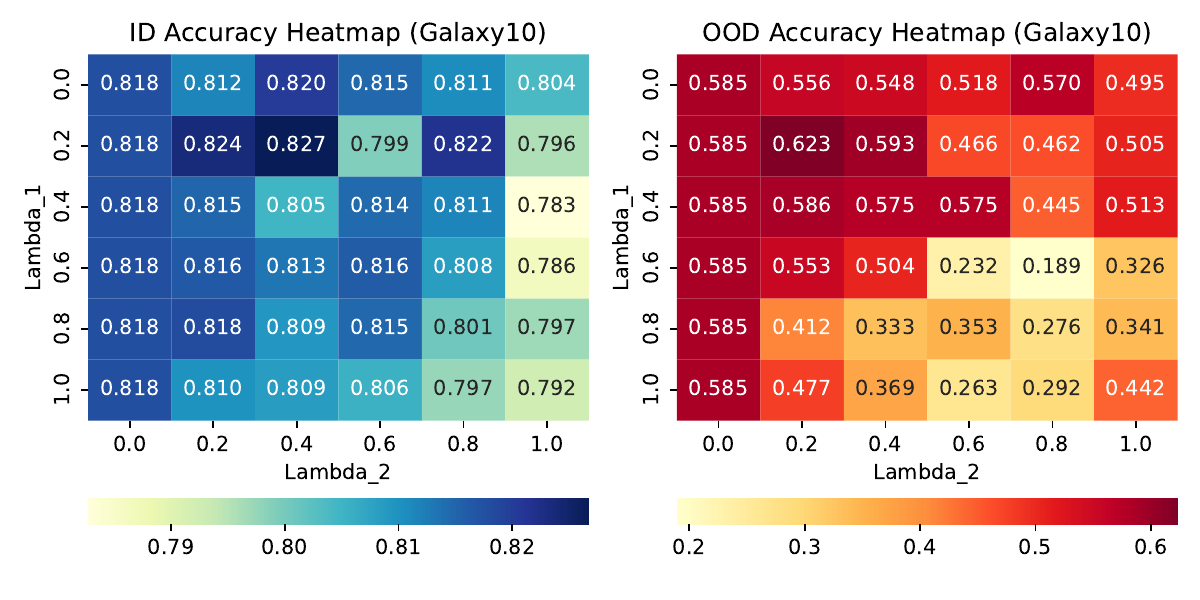}}
    \subfigure[F1 Score Heatmaps on BirdCalls.]{\includegraphics[width=0.5\linewidth]{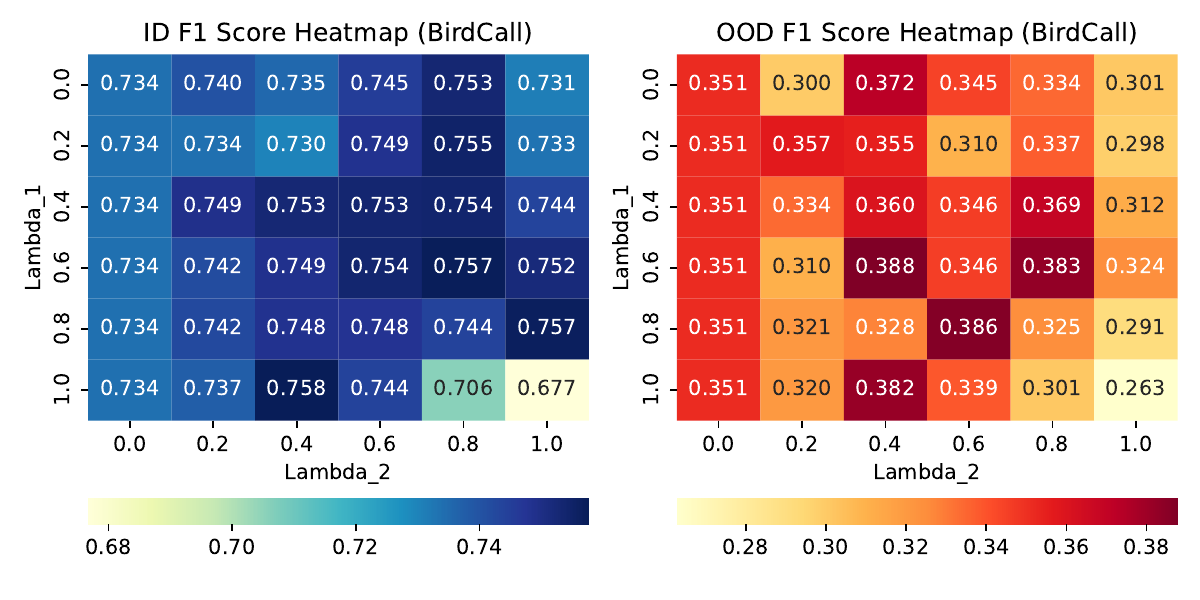}}
  \\
  \caption{Effect of Mixing Ratios on ID and OOD Performance.}
  \label{fig:mixing-heatmap}
\end{figure}

To further understand the effect of mixing parameters, we visualize ID and OOD performance under different values of pixel-space blending ratio $\lambda_1$ and final fusion ratio $\lambda_2$ in Figure~\ref{fig:mixing-heatmap}. We conduct this analysis on Galaxy10 and BirdCalls. The results show that ID performance is relatively stable across different combinations, while OOD performance shows obvious variation. 

A smaller $\lambda_1$ preserves more pixel-level details while a larger one introduces a stronger distribution shift of the augmented image. The second-stage fusion parameter $\lambda_2$ then determines how much of the pixel-augmented image is preserved versus the frequency-based variant that randomizes the domain-dependent features.
We observe a key pattern: when $\lambda_1$ is large (i.e., the image already deviates strongly from the source domain), increasing $\lambda_2$ further degrades OOD performance, because it introduces excessive deviation, losing the semantic identity of the source image. In contrast, when $\lambda_1$ is small (i.e., the pixel-space mix preserves the original details well), a larger $\lambda_2$ does not degrade the OOD performance as intensively as a smaller $\lambda_2$. Overall, moderate $\lambda_1$ and $\lambda_2$ tend to result in better OOD performance.

These results highlight the importance of balancing the two stages: $\lambda_1$ should introduce moderate domain shift while keeping original pixel-level details, and $\lambda_2$ should complement it by preserving the semantic information. From a feature decomposition perspective, balanced $\lambda_1$ and $\lambda_2$ perturbs domain-specific spurious factors ($x_{d\text{:spu}}$) while preserving label-relevant features ($x_{\text{obj}}, x_{d\text{:robust}}$). Over-mixing in each stage, however, risks disrupting $x_{\text{obj}}$ or a lack of $x_{d\text{:spu}}$ perturbation, ultimately degrading OOD performance.

\section{Conclusion, Limitations and Future Work}
In this work, we propose Frequency-Pixel Connect, which unifies frequency and pixel information in targeted augmentation to improve out-of-distribution (OOD) robustness. By jointly perturbing domain-dependent features while preserving semantic structures, our method achieves consistent improvements across multiple real-world datasets.

However, we acknowledge a limitation: OOD performance of our method still depends on the proper tuning of the mixing ratio $\lambda_1$ and $\lambda_2$. Despite this, the trade-off is acceptable in most domain adaptation scenarios, and our approach remains significantly more flexible and easier to apply compared to methods that require extensive domain-specific insights.

In future work, we aim to extend this framework to more complex scenarios and investigate adaptive strategies that learn the mixing parameters automatically. Moreover, integrating our augmentation scheme with foundation models or self-supervised objectives may further improve OOD robustness in low-label or zero-shot settings.

\bibliography{ref}

\begin{thebibliography}{10}

\bibitem{bandi2018detection}
Peter Bandi, Oscar Geessink, Quirine Manson, Marcory Van~Dijk, Maschenka Balkenhol, Meyke Hermsen, Babak~Ehteshami Bejnordi, Byungjae Lee, Kyunghyun Paeng, Aoxiao Zhong, et~al.
\newblock From detection of individual metastases to classification of lymph node status at the patient level: the camelyon17 challenge.
\newblock {\em IEEE transactions on medical imaging}, 38(2):550--560, 2018.

\bibitem{beery2021iwildcam}
Sara Beery, Arushi Agarwal, Elijah Cole, and Vighnesh Birodkar.
\newblock The iwildcam 2021 competition dataset.
\newblock {\em arXiv preprint arXiv:2105.03494}, 2021.

\bibitem{beery2018recognition}
Sara Beery, Grant Van~Horn, and Pietro Perona.
\newblock Recognition in terra incognita.
\newblock In {\em Proceedings of the European conference on computer vision (ECCV)}, pages 456--473, 2018.

\bibitem{caron2020unsupervised}
Mathilde Caron, Ishan Misra, Julien Mairal, Priya Goyal, Piotr Bojanowski, and Armand Joulin.
\newblock Unsupervised learning of visual features by contrasting cluster assignments.
\newblock {\em Advances in neural information processing systems}, 33:9912--9924, 2020.

\bibitem{chen2020simple}
Ting Chen, Simon Kornblith, Mohammad Norouzi, and Geoffrey Hinton.
\newblock A simple framework for contrastive learning of visual representations.
\newblock In {\em International conference on machine learning}, pages 1597--1607. PmLR, 2020.

\bibitem{cubuk2020randaugment}
Ekin~D Cubuk, Barret Zoph, Jonathon Shlens, and Quoc~V Le.
\newblock Randaugment: Practical automated data augmentation with a reduced search space.
\newblock In {\em Proceedings of the IEEE/CVF conference on computer vision and pattern recognition workshops}, pages 702--703, 2020.

\bibitem{devries2017improved}
Terrance DeVries and Graham~W Taylor.
\newblock Improved regularization of convolutional neural networks with cutout.
\newblock {\em arXiv preprint arXiv:1708.04552}, 2017.

\bibitem{ganin2016domain}
Yaroslav Ganin, Evgeniya Ustinova, Hana Ajakan, Pascal Germain, Hugo Larochelle, Fran{\c{c}}ois Laviolette, Mario March, and Victor Lempitsky.
\newblock Domain-adversarial training of neural networks.
\newblock {\em Journal of machine learning research}, 17(59):1--35, 2016.

\bibitem{gao2023out}
Irena Gao, Shiori Sagawa, Pang~Wei Koh, Tatsunori Hashimoto, and Percy Liang.
\newblock Out-of-domain robustness via targeted augmentations.
\newblock In {\em International Conference on Machine Learning}, pages 10800--10834. PMLR, 2023.

\bibitem{gulrajanisearch}
Ishaan Gulrajani and David Lopez-Paz.
\newblock In search of lost domain generalization.
\newblock In {\em International Conference on Learning Representations}, 2021.

\bibitem{hansen2007structural12}
Bruce~C Hansen and Robert~F Hess.
\newblock Structural sparseness and spatial phase alignment in natural scenes.
\newblock {\em Journal of the Optical Society of America A}, 24(7):1873--1885, 2007.

\bibitem{he2016deep}
Kaiming He, Xiangyu Zhang, Shaoqing Ren, and Jian Sun.
\newblock Deep residual learning for image recognition.
\newblock In {\em Proceedings of the IEEE conference on computer vision and pattern recognition}, pages 770--778, 2016.

\bibitem{hendrycks2021many}
Dan Hendrycks, Steven Basart, Norman Mu, Saurav Kadavath, Frank Wang, Evan Dorundo, Rahul Desai, Tyler Zhu, Samyak Parajuli, Mike Guo, et~al.
\newblock The many faces of robustness: A critical analysis of out-of-distribution generalization.
\newblock In {\em Proceedings of the IEEE/CVF international conference on computer vision}, pages 8340--8349, 2021.

\bibitem{hendrycksaugmix}
Dan Hendrycks, Norman Mu, Ekin~Dogus Cubuk, Barret Zoph, Justin Gilmer, and Balaji Lakshminarayanan.
\newblock Augmix: A simple data processing method to improve robustness and uncertainty.
\newblock In {\em International Conference on Learning Representations}, 2020.

\bibitem{Galaxy10}
Leung Henry.
\newblock Galaxy10 decals dataset.
\newblock https://github.com/henrysky/Galaxy10, 2021.

\bibitem{hoffman2018cycada}
Judy Hoffman, Eric Tzeng, Taesung Park, Jun-Yan Zhu, Phillip Isola, Kate Saenko, Alexei Efros, and Trevor Darrell.
\newblock Cycada: Cycle-consistent adversarial domain adaptation.
\newblock In {\em International conference on machine learning}, pages 1989--1998. Pmlr, 2018.

\bibitem{hoffman2018cycada16}
Judy Hoffman, Eric Tzeng, Taesung Park, Jun-Yan Zhu, Phillip Isola, Kate Saenko, Alexei Efros, and Trevor Darrell.
\newblock Cycada: Cycle-consistent adversarial domain adaptation.
\newblock In {\em International conference on machine learning}, pages 1989--1998. Pmlr, 2018.

\bibitem{ilse2021selecting}
Maximilian Ilse, Jakub~M Tomczak, and Patrick Forr{\'e}.
\newblock Selecting data augmentation for simulating interventions.
\newblock In {\em International conference on machine learning}, pages 4555--4562. PMLR, 2021.

\bibitem{joly2022overview}
Alexis Joly, Herv{\'e} Go{\"e}au, Stefan Kahl, Luk{\'a}{\v{s}} Picek, Titouan Lorieul, Elijah Cole, Benjamin Deneu, Maximilien Servajean, Andrew Durso, Herv{\'e} Glotin, et~al.
\newblock Overview of lifeclef 2022: an evaluation of machine-learning based species identification and species distribution prediction.
\newblock In {\em International Conference of the Cross-Language Evaluation Forum for European Languages}, pages 257--285. Springer, 2022.

\bibitem{kang2019contrastive}
Guoliang Kang, Lu~Jiang, Yi~Yang, and Alexander~G Hauptmann.
\newblock Contrastive adaptation network for unsupervised domain adaptation.
\newblock In {\em Proceedings of the IEEE/CVF conference on computer vision and pattern recognition}, pages 4893--4902, 2019.

\bibitem{koh2021wilds}
Pang~Wei Koh, Shiori Sagawa, Henrik Marklund, Sang~Michael Xie, Marvin Zhang, Akshay Balsubramani, Weihua Hu, Michihiro Yasunaga, Richard~Lanas Phillips, Irena Gao, et~al.
\newblock Wilds: A benchmark of in-the-wild distribution shifts.
\newblock In {\em International conference on machine learning}, pages 5637--5664. PMLR, 2021.

\bibitem{krizhevsky2012imagenet}
Alex Krizhevsky, Ilya Sutskever, and Geoffrey~E Hinton.
\newblock Imagenet classification with deep convolutional neural networks.
\newblock {\em Advances in neural information processing systems}, 25, 2012.

\bibitem{kumar2022fine}
Ananya Kumar, Aditi Raghunathan, Robbie Jones, Tengyu Ma, and Percy Liang.
\newblock Fine-tuning can distort pretrained features and underperform out-of-distribution.
\newblock In {\em International Conference on Learning Representations}, 2022.

\bibitem{long2018conditional}
Mingsheng Long, Zhangjie Cao, Jianmin Wang, and Michael~I Jordan.
\newblock Conditional adversarial domain adaptation.
\newblock {\em Advances in neural information processing systems}, 31, 2018.

\bibitem{ng2020ssmba}
Nathan Ng, Kyunghyun Cho, and Marzyeh Ghassemi.
\newblock Ssmba: Self-supervised manifold based data augmentation for improving out-of-domain robustness.
\newblock {\em arXiv preprint arXiv:2009.10195}, 2020.

\bibitem{oppenheim1979phase33}
A~Oppenheim, Jae Lim, Gary Kopec, and SC~Pohlig.
\newblock Phase in speech and pictures.
\newblock In {\em ICASSP'79. IEEE International Conference on Acoustics, Speech, and Signal Processing}, volume~4, pages 632--637. IEEE, 1979.

\bibitem{oppenheim1981importance34}
Alan~V Oppenheim and Jae~S Lim.
\newblock The importance of phase in signals.
\newblock {\em Proceedings of the IEEE}, 69(5):529--541, 1981.

\bibitem{park2019specaugment}
Daniel~S Park, William Chan, Yu~Zhang, Chung-Cheng Chiu, Barret Zoph, Ekin~D Cubuk, and Quoc~V Le.
\newblock Specaugment: A simple data augmentation method for automatic speech recognition.
\newblock {\em arXiv preprint arXiv:1904.08779}, 2019.

\bibitem{piotrowski1982demonstration36}
Leon~N Piotrowski and Fergus~W Campbell.
\newblock A demonstration of the visual importance and flexibility of spatial-frequency amplitude and phase.
\newblock {\em Perception}, 11(3):337--346, 1982.

\bibitem{qu2024connect}
Helen Qu and Sang~Michael Xie.
\newblock Connect later: improving fine-tuning for robustness with targeted augmentations.
\newblock In {\em Proceedings of the 41st International Conference on Machine Learning}, pages 41769--41786, 2024.

\bibitem{radford2021learning}
Alec Radford, Jong~Wook Kim, Chris Hallacy, Aditya Ramesh, Gabriel Goh, Sandhini Agarwal, Girish Sastry, Amanda Askell, Pamela Mishkin, Jack Clark, et~al.
\newblock Learning transferable visual models from natural language supervision.
\newblock In {\em International conference on machine learning}, pages 8748--8763. PmLR, 2021.

\bibitem{rebuffi2021data}
Sylvestre-Alvise Rebuffi, Sven Gowal, Dan~Andrei Calian, Florian Stimberg, Olivia Wiles, and Timothy~A Mann.
\newblock Data augmentation can improve robustness.
\newblock {\em Advances in neural information processing systems}, 34:29935--29948, 2021.

\bibitem{sagawaextending}
Shiori Sagawa, Pang~Wei Koh, Tony Lee, Irena Gao, Sang~Michael Xie, Kendrick Shen, Ananya Kumar, Weihua Hu, Michihiro Yasunaga, Henrik Marklund, et~al.
\newblock Extending the wilds benchmark for unsupervised adaptation.
\newblock In {\em International Conference on Learning Representations}, 2022.

\bibitem{sainburg2024noisereduce}
Tim Sainburg and Asaf Zorea.
\newblock Noisereduce: Domain general noise reduction for time series signals.
\newblock {\em arXiv preprint arXiv:2412.17851}, 2024.

\bibitem{sarkar2024demystifying}
Ankita Sarkar, Sarbani Palit, and Ujjwal Bhattacharya.
\newblock Demystifying galaxy classification: An elegant and powerful hybrid approach.
\newblock In {\em 2024 39th International Conference on Image and Vision Computing New Zealand (IVCNZ)}, pages 1--6. IEEE, 2024.

\bibitem{shen2022connect}
Kendrick Shen, Robbie~M Jones, Ananya Kumar, Sang~Michael Xie, Jeff~Z HaoChen, Tengyu Ma, and Percy Liang.
\newblock Connect, not collapse: Explaining contrastive learning for unsupervised domain adaptation.
\newblock In {\em International conference on machine learning}, pages 19847--19878. PMLR, 2022.

\bibitem{simard2003best}
Patrice~Y Simard, David Steinkraus, John~C Platt, et~al.
\newblock Best practices for convolutional neural networks applied to visual document analysis.
\newblock In {\em Icdar}, volume~3. Edinburgh, 2003.

\bibitem{sohn2020fixmatch}
Kihyuk Sohn, David Berthelot, Nicholas Carlini, Zizhao Zhang, Han Zhang, Colin~A Raffel, Ekin~Dogus Cubuk, Alexey Kurakin, and Chun-Liang Li.
\newblock Fixmatch: Simplifying semi-supervised learning with consistency and confidence.
\newblock {\em Advances in neural information processing systems}, 33:596--608, 2020.

\bibitem{sun2017correlation}
Baochen Sun, Jiashi Feng, and Kate Saenko.
\newblock Correlation alignment for unsupervised domain adaptation.
\newblock {\em Domain adaptation in computer vision applications}, pages 153--171, 2017.

\bibitem{sun2016deep}
Baochen Sun and Kate Saenko.
\newblock Deep coral: Correlation alignment for deep domain adaptation.
\newblock In {\em Computer vision--ECCV 2016 workshops: Amsterdam, the Netherlands, October 8-10 and 15-16, 2016, proceedings, part III 14}, pages 443--450. Springer, 2016.

\bibitem{tellez2018whole}
David Tellez, Maschenka Balkenhol, Irene Otte-H{\"o}ller, Rob Van De~Loo, Rob Vogels, Peter Bult, Carla Wauters, Willem Vreuls, Suzanne Mol, Nico Karssemeijer, et~al.
\newblock Whole-slide mitosis detection in h\&e breast histology using phh3 as a reference to train distilled stain-invariant convolutional networks.
\newblock {\em IEEE transactions on medical imaging}, 37(9):2126--2136, 2018.

\bibitem{wang2023galaxy}
Guangze Wang.
\newblock Galaxy morphology classification with densenet.
\newblock In {\em Journal of Physics: Conference Series}, volume 2580, page 012064. IOP Publishing, 2023.

\bibitem{wilesfine}
Olivia Wiles, Sven Gowal, Florian Stimberg, Sylvestre-Alvise Rebuffi, Ira Ktena, Krishnamurthy~Dj Dvijotham, and Ali~Taylan Cemgil.
\newblock A fine-grained analysis on distribution shift.
\newblock In {\em International Conference on Learning Representations}, 2021.

\bibitem{xie2020unsuperviseda}
Qizhe Xie, Zihang Dai, Eduard Hovy, Thang Luong, and Quoc Le.
\newblock Unsupervised data augmentation for consistency training.
\newblock {\em Advances in neural information processing systems}, 33:6256--6268, 2020.

\bibitem{xie2020self}
Qizhe Xie, Minh-Thang Luong, Eduard Hovy, and Quoc~V Le.
\newblock Self-training with noisy student improves imagenet classification.
\newblock In {\em Proceedings of the IEEE/CVF conference on computer vision and pattern recognition}, pages 10687--10698, 2020.

\bibitem{xu2023fourier}
Qinwei Xu, Ruipeng Zhang, Ziqing Fan, Yanfeng Wang, Yi-Yan Wu, and Ya~Zhang.
\newblock Fourier-based augmentation with applications to domain generalization.
\newblock {\em Pattern Recognition}, 139:109474, 2023.

\bibitem{yan2020improve}
Shen Yan, Huan Song, Nanxiang Li, Lincan Zou, and Liu Ren.
\newblock Improve unsupervised domain adaptation with mixup training.
\newblock {\em arXiv preprint arXiv:2001.00677}, 2020.

\bibitem{yang2021interactive}
Qize Yang, Xihan Wei, Biao Wang, Xian-Sheng Hua, and Lei Zhang.
\newblock Interactive self-training with mean teachers for semi-supervised object detection.
\newblock In {\em Proceedings of the IEEE/CVF conference on computer vision and pattern recognition}, pages 5941--5950, 2021.

\bibitem{yang2020phase48}
Yanchao Yang, Dong Lao, Ganesh Sundaramoorthi, and Stefano Soatto.
\newblock Phase consistent ecological domain adaptation.
\newblock In {\em Proceedings of the IEEE/CVF conference on computer vision and pattern recognition}, pages 9011--9020, 2020.

\bibitem{yang2020fda49}
Yanchao Yang and Stefano Soatto.
\newblock Fda: Fourier domain adaptation for semantic segmentation.
\newblock In {\em Proceedings of the IEEE/CVF conference on computer vision and pattern recognition}, pages 4085--4095, 2020.

\bibitem{yao2022improving}
Huaxiu Yao, Yu~Wang, Sai Li, Linjun Zhang, Weixin Liang, James Zou, and Chelsea Finn.
\newblock Improving out-of-distribution robustness via selective augmentation.
\newblock In {\em International Conference on Machine Learning}, pages 25407--25437. PMLR, 2022.

\bibitem{yun2019cutmix}
Sangdoo Yun, Dongyoon Han, Seong~Joon Oh, Sanghyuk Chun, Junsuk Choe, and Youngjoon Yoo.
\newblock Cutmix: Regularization strategy to train strong classifiers with localizable features.
\newblock In {\em Proceedings of the IEEE/CVF international conference on computer vision}, pages 6023--6032, 2019.

\bibitem{zhang2017mixup}
Hongyi Zhang, Moustapha Cisse, Yann~N Dauphin, and David Lopez-Paz.
\newblock mixup: Beyond empirical risk minimization.
\newblock {\em arXiv preprint arXiv:1710.09412}, 2017.

\bibitem{zhou2020deep}
Kaiyang Zhou, Yongxin Yang, Timothy Hospedales, and Tao Xiang.
\newblock Deep domain-adversarial image generation for domain generalisation.
\newblock In {\em Proceedings of the AAAI conference on artificial intelligence}, volume~34, pages 13025--13032, 2020.

\end{thebibliography}

\clearpage

\appendix

\section{Related Work}
\subsection{Pretraining for OOD Robustness} 
Pretraining on large-scale unlabeled data has emerged as a promising OOD generalization method \cite{caron2020unsupervised, shen2022connect, radford2021learning, sagawaextending}. Contrastive pretraining, which encourages feature alignment across domains using generic augmentations, outperforms traditional domain adaptation methods that enforce explicit domain invariance \cite{kang2019contrastive, hoffman2018cycada}. However, the standard fine-tuning with generic augmentations often leads to unsatisfying OOD performance due to misalignment with specific distribution shifts \cite{qu2024connect, kumar2022fine}. Although Connect Later \cite{qu2024connect} improves OOD robustness by applying targeted augmentations during linear probing then fine-tuning (LP-FT) \cite{kumar2022fine}, it still relies heavily on manually designed augmentations and cannot easily generalize without shift-specific analysis. 

\subsection{Augmentations}
For a given input $x \in \mathcal{X}$, augmented variants $x' \sim \mathcal{A}(\cdot \mid x)$ are often used to enhance robustness \cite{hendrycksaugmix, hendrycks2021many}. 
Data augmentation has been widely used to improve model robustness to label-invariant transformations such as translation or rotation \cite{hendrycksaugmix, rebuffi2021data, ng2020ssmba}. Traditional augmentation strategies typically apply generic perturbations to increase input diversity \citep{simard2003best, krizhevsky2012imagenet, cubuk2020randaugment, devries2017improved, zhang2017mixup}. However, recent studies highlight that the choice of augmentation type significantly impacts model generalization \citep{chen2020simple, xie2020unsuperviseda}.

In semi-supervised learning, augmentations have also been utilized in the self-training paradigm to improve generalization by training on pseudo-labeled data \cite{xie2020self, sohn2020fixmatch, yang2021interactive}. Nevertheless, Qu et al. \cite{qu2024connect} show that a self-training baseline using pseudo-labels generated from an ERM model with targeted augmentations still underperforms compared to Connect Later \cite{qu2024connect}. This underscores the importance of pretrained representations in achieving robustness gains. Connect Later \cite{qu2024connect} leverages targeted augmentations not only as a data transformation tool, but as a design interface to incorporate knowledge of the distribution shift while still benefiting from pretrained features.

Recent work in domain shift problems emphasizes the effectiveness of \textit{targeted augmentations}—those informed by the nature of the distribution shift—over generic or even target-aware augmentations such as CutMix \cite{gao2023out, yun2019cutmix}. While \cite{gao2023out} focus on domain generalization where the target domain is unknown, some recent work \cite{qu2024connect,gao2023out} studies the domain adaptation setting with access to unlabeled target data. The targeted augmentation is effective but rely on manually designing and are dataset-specialized. 

\subsection{On the Importance of Phase Information}
A series of early works \cite{oppenheim1979phase33, oppenheim1981importance34, piotrowski1982demonstration36,hansen2007structural12} have highlighted that in the Fourier spectrum of visual signals, the phase component is critical for preserving high-level semantics, while the amplitude primarily captures low-level appearance statistics. Recent advances further explore this perspective in domain adaptation. For example, \cite{yang2020fda49} propose to replace a local region in the centralized amplitude spectrum of a source image with that of a target image to synthesize target-style images for training. In parallel, \cite{yang2020phase48} introduce phase consistency as a constraint during source-target translation, demonstrating its superiority over traditional cycle consistency \cite{hoffman2018cycada16} in domain adaptation tasks, especially for semantic segmentation. Xu et al. \cite{xu2023fourier} incorporates a Fourier-based strategy that emphasizes phase preservation to enhance generalization performance under domain shift. Inspired by this, we propose to use phase in frequency space to preserve the task-relevant features in our augmentations.

\section{Dataset Details}
\label{Dataset Details}
\paragraph{Species Classification from Camera Trap Images (iWildCam).} 
In the iWildCam dataset \cite{beery2021iwildcam, koh2021wilds}, the task is to classify an animal species \(y\) from images \(x\) captured by static camera traps. The dataset consists of 243 camera locations in \(D^{train}\). Although images from the same camera share similar backgrounds, domain-dependent features \(x_{d: robust}\), such as habitat, vary across domains. For example, cameras in Kenyan savannas (cameras 23 and 97) and a Guatemalan forest (camera 54) have differing background features. This requires the model to focus on foreground features (e.g., animal species) for accurate prediction, as background habitat often doesn't provide useful information for classification \cite{gao2023out, qu2024connect}.

\paragraph{Tumor Identification in Histopathology Slides (Camelyon17).}
In Camelyon17 \cite{beery2021iwildcam, koh2021wilds}, the task is to classify tumor presence in histopathology slides. These slides come from three hospitals, each with varying imaging protocols and staining techniques that lead to domain-specific color variations. For example, some hospitals have mostly early-stage pN1 breast cancer, while others have later-stage pN2 cases. These domain-specific features, such as tumor size and density, correlate with the cancer stage \(y\), requiring the model to account for these domain shifts in order to generalize effectively \cite{gao2023out, qu2024connect}.

\paragraph{Bird Species Recognition from Audio Recordings (BirdCalls).}
The BirdCalls dataset \cite{joly2022overview, gao2023out} consists of audio recordings from nine microphones located in different regions. The task is to classify the bird species \(y\) vocalizing in each audio clip \(x\). While the bird species information is crucial, low-level noise and microphone settings (e.g., gain levels) often introduce irrelevant domain-specific features. These features, such as background noises from different habitats, should be ignored by the model, which should focus on bird calls for accurate classification. Randomizing microphone-related features ensures that the model generalizes across diverse recording environments \cite{gao2023out}.

\paragraph{Astronomical Morphology Classification (Galaxy10 DECaLS\&SDSS).}
Galaxy 10 DECaLS and SDSS are two galaxy morphology classification datasets containing images from two different surveys \cite{Galaxy10}. They differ in image quality, resolution, and color distribution due to differences in telescope instrumentation. We treat Galaxy10-DECaLS as the source domain and Galaxy10-SDSS as the target domain, and evaluate cross-domain generalization without using any target-domain labels during training.

\section{Experimental Details}
\label{Experimental Details}

\subsection{Frequency-Pixel Connect}
We listed some important implementation details of our method as follows. Others details can be find in our code in the supplementary material.

\paragraph{iWildCam.}
We use ResNet-50 pretrained with SwAV \cite{caron2020unsupervised} as the backbone \cite{qu2024connect}. For pretrained models, we follow LP-FT \cite{kumar2022fine, qu2024connect}: a linear probe is trained for 10 epochs using Adam, and its weights are used to initialize the full model, which is then fine-tuned for another 20 epochs. Following \cite{qu2024connect}, we perform 10 hyperparameter search trials by sampling: linear probe learning rate $10^{\text{Uniform}[-3,-2]}$, fine-tuning learning rate $10^{\text{Uniform}[-5,-2]}$, and augmentation probability from $\text{Uniform}[0.5, 0.9]$. The configuration with the best OOD validation performance is selected for testing. To mitigate outlier variance, our method is averaged over 5 seeds. Othor hyperparameters are the same as that of Qu et al. \cite{qu2024connect}.

\paragraph{Camelyon17.}
We conduct experiments on the unlabeled Camelyon17 dataset \cite{sagawaextending} using DenseNet121 pretrained with SwAV \cite{caron2020unsupervised}, following the protocol in \cite{qu2024connect, gao2023out}. For pretrained models, we first apply 10 epochs of linear probing before full fine-tuning for 20 epochs, and select the linear probe checkpoint with the highest OOD validation accuracy for initialization. For hyperparameter tuning, following \cite{qu2024connect}, the following hyperparameters are sampled: linear probe learning rate from $10^{\text{Uniform}[-3,-2]}$, fine-tuning learning rate from $10^{\text{Uniform}[-5,-2]}$, augmentation probability from $\text{Uniform}[0.5, 0.9]$, and augmentation strength from $\text{Uniform}[0.05, 0.1]$. The best hyperparameter configuration is selected based on OOD validation performance. All results are averaged over 5 random seeds.  Othor hyperparameters are the same as that of Qu et al. \cite{qu2024connect}.

\paragraph{BirdCalls.}
We evaluate our method on the BirdCalls dataset using a pretrained EfficientNet-B0 encoder. Following \cite{gao2023out}, we first perform 10 epochs of linear probing (LP-FT \cite{kumar2022fine}) with the Adam optimizer, selecting the checkpoint with the best OOD validation performance to initialize the fine-tuning stage for 30 epochs.
Consistent with the protocol in \cite{qu2024connect}, we sample the linear probe learning rate from $10^{\text{Uniform}[-3,-2]}$, the fine-tuning learning rate from $10^{\text{Uniform}[-4,-3]}$, the augmentation probability from $\text{Uniform}[0.5, 0.9]$. The final hyperparameter setting is selected based on the best OOD validation performance. Reported results are averaged over 5 random seeds.  Othor hyperparameters are the same as that of Gao et al. \cite{gao2023out}.

\paragraph{Galaxy10.}
We evaluate our method on the Galaxy10 DECaLS and Galaxy10 SDSS datasets. Specifically, we use the DECaLS for training and validation, and evaluate the model on the SDSS subset, where labels are mapped to DECaLS categories using a predefined label conversion rule since the labels in the two datasets are not exactly the same. 
Our training pipeline employs a ResNet18 encoder initialized from a contrastively pre-trained checkpoint, followed by a linear classifier trained end-to-end. We first perform 10 epochs of linear probing (LP-FT \cite{kumar2022fine}) with the Adam optimizer, selecting the checkpoint with the best OOD validation performance to initialize the fine-tuning stage. The fine-tuning stage lasts for 15 epochs using Adam optimizer with a learning rate of $10^{-4}$ and a batch size of 8. The augmentation probability is set to 0.5.

In Frequency-Pixel Mixing, the target image for mixing can be selected from either labeled or unlabeled data. For consistency and fair comparison with baseline methods \cite{gao2023out, qu2024connect}, we adopt a unified strategy across datasets. Specifically, for iWildCam, Camelyon17, and BirdCalls, we mix each source image with samples from other domains within the training set. For Galaxy10, since the training data only contains images from a single telescope, we randomly sample images from the unlabeled test set as mixing targets to generate cross-domain augmented images.

\subsection{Baseline Methods}
For baseline methods on iWildCam, Camelyon17, BirdCalls dataset, we followed the settings and hyperparameters exactly in \cite{qu2024connect,gao2023out}. And for galaxy 10 dataset, the settings are presented in Table \ref{settings}.

\begin{table}[t]
\centering
\caption{Hyperparameter settings for baseline methods on Galaxy10.}
\label{settings}
\begin{tabular}{ll}
\toprule
\textbf{Method} & \textbf{Hyperparameters} \\
\midrule
ERM & \begin{tabular}[t]{@{}l@{}}
Learning rate $= 10^{-4}$ \\
Weight decay $= 0$
\end{tabular} \\
\midrule
LISA & \begin{tabular}[t]{@{}l@{}}
Learning rate $= 10^{-4}$ \\
Weight decay $= 0$ \\
Transform probability $= 0.9$
\end{tabular} \\
\midrule
MixUp & \begin{tabular}[t]{@{}l@{}}
Learning rate $= 10^{-4}$ \\
Weight decay $= 0$ \\
Transform probability $= 0.9$ \\
$\alpha = 0.4$
\end{tabular} \\
\midrule
CutMix & \begin{tabular}[t]{@{}l@{}}
Learning rate $= 10^{-4}$ \\
Weight decay $= 0$ \\
Transform probability $= 0.9$ \\
$\alpha = 0.4$
\end{tabular} \\
\midrule
Cutout & \begin{tabular}[t]{@{}l@{}}
Learning rate $= 10^{-4}$ \\
Weight decay $= 0$ \\
Transform probability $= 0.9$ \\
$\alpha = 0.4$
\end{tabular} \\
\midrule
Randaugment & \begin{tabular}[t]{@{}l@{}}
Learning rate $= 10^{-4}$ \\
Transform probability $= 0.9$ \\
Version = $Original$
\end{tabular} \\
\midrule
CORAL & \begin{tabular}[t]{@{}l@{}}
Learning rate $= 10^{-4}$ \\
Coral weight $= 1.0$ 
\end{tabular} \\
\midrule
Noisy Student & \begin{tabular}[t]{@{}l@{}}
Learning rate $= 10^{-4}$ \\
soft label $=False$
\end{tabular} \\
\midrule
DANN & \begin{tabular}[t]{@{}l@{}}
Classifier learning rate $= 10^{-4}$ \\
Discriminator learning rate $= 10^{-4}$ \\
GRL$\lambda = 2 / (1 + exp(-10·epoch/15)) - 1$
\end{tabular} \\
\bottomrule
\end{tabular}
\end{table}

\section{Computation of Connectivity}
\label{appendix_connectivity}
To compute the average connectivity between two class-domain pairs $(c, d)$ and $(c', d')$, the process involves labeling all training examples from class $c$ and domain $d$ as 0, and those from class $c'$ and domain $d'$ as 1, while discarding examples from other classes or domains. 
Then, a ResNet50 model is trained using strong augmentations, stochastic gradient descent (SGD) with momentum, and a cosine learning rate schedule, maintaining consistent training parameters across all classes and domains. We train using the same hyperparameters described in \ref{Experimental Details} for 3,000 steps, applying early stopping based on validation accuracy.

\end{document}